\documentclass[conference]{IEEEtran}
\usepackage{float}
\IEEEoverridecommandlockouts
\usepackage{cite}
\usepackage{amsmath,amssymb,amsfonts}
\usepackage{algorithmic}
\usepackage{graphicx}

\usepackage[shortcuts,acronym, nopostdot,style=super,nonumberlist,toc]{glossaries}

\usepackage{hyperref}
\usepackage{textcomp}
\usepackage{xcolor}
\usepackage{multirow}

\usepackage{lipsum}

\def\BibTeX{{\rm B\kern-.05em{\sc i\kern-.025em b}\kern-.08em
    T\kern-.1667em\lower.7ex\hbox{E}\kern-.125emX}}

\makeglossaries
\newacronym{AI}{AI}{Artificial Intelligence}
\newacronym{ML}{ML}{Machine Learning}
\newacronym{EU}{EU}{European Union}
\newacronym{AIA}{Act}{EU AI Act}
\newacronym{TSR}{TSR}{Traffic Sign Recognition}
\newacronym{OEM}{OEM}{Original Equipment Manufacturer}

\usepackage{tikz}
\newcommand\copyrighttext{%
  \footnotesize \textcopyright 2024 IEEE. Personal use of this material is permitted.
  Permission from IEEE must be obtained for all other uses, in any current or future
  media, including reprinting/republishing this material for advertising or promotional
  purposes, creating new collective works, for resale or redistribution to servers or
  lists, or reuse of any copyrighted component of this work in other works.}
\newcommand\copyrightnotice{%
\begin{tikzpicture}[remember picture,overlay]
\node[anchor=south,yshift=10pt] at (current page.south) {\fbox{\parbox{\dimexpr\textwidth-\fboxsep-\fboxrule\relax}{\copyrighttext}}};
\end{tikzpicture}%
}

\begin{document}
\title{Navigating the EU AI Act: A Methodological Approach to Compliance for Safety-critical Products  \\
}

\author{\IEEEauthorblockN{Jessica Kelly, Shanza Ali Zafar, Lena Heidemann, Jo\~{a}o-Vitor Zacchi, Delfina Espinoza, N\'{u}ria Mata}
\IEEEauthorblockA{
\textit{Fraunhofer IKS}\\
Munich, Germany \\
jessica.kelly@iks.fraunhofer.de}
}

\maketitle
\copyrightnotice

\begin{abstract}
In December 2023, the European Parliament provisionally agreed on the EU AI Act. This unprecedented regulatory framework for AI systems lays out guidelines to ensure the safety, legality, and trustworthiness of AI products. This paper presents a methodology for interpreting the EU AI Act requirements for high-risk AI systems by leveraging product quality models. We first propose an extended product quality model for AI systems, incorporating attributes relevant to the Act not covered by current quality models. We map the Act requirements to relevant quality attributes with the goal of refining them into measurable characteristics.  We then propose a contract-based approach to derive technical requirements at the stakeholder level. This facilitates the development and assessment of \ac{AI} systems that not only adhere to established quality standards, but also comply with the regulatory requirements outlined in the Act for high-risk (including safety-critical) \ac{AI} systems. We demonstrate the applicability of this methodology on an exemplary automotive supply chain use case, where several stakeholders interact to achieve EU AI Act compliance. 
\end{abstract}


\section{Introduction}
With the growing number of \ac{AI} systems being deployed in safety-critical applications, there is a pressing need to establish regulations that govern the safe and responsible use of \ac{AI}. In December 2023, the \ac{AIA} was provisionally agreed upon by the European Parliament, setting the precedent for the regulation of \ac{AI} applications. It is the first comprehensive regulatory framework governing the development, deployment, and use of \ac{AI} systems. The \ac{AIA} introduces a risk-based classification of \ac{AI} products. Applications whose risk is deemed ``Unacceptable'', such as social-scoring systems, are banned within the framework of the \ac{AIA}. Applications with a risk rated ``High'' (high-risk) must demonstrate compliance with stringent requirements ensuring that, among others, safety, transparency, and human rights needs are met. The outlined requirements affect not only \ac{AI} products, but any stakeholders involved in the \ac{AI} value chain. Organizations will need to adapt to the evolving landscape of the \ac{AIA}, balancing innovation and regulatory adherence. While entities across the \ac{AI} value chain will need to align with the framework of the \ac{AIA}, future standards and regulations will also be affected.

Safety-critical systems, whose failure could result in significant harm to people or even loss of life, fit under the definition of high risk as defined by the \ac{AIA}. Existing domain-specific safety standards, such as ISO 26262 \cite{iso26262} for automotive or ARP 4761 \cite{arp4761}, DO-178C \cite{do178c}, and DO-254 \cite{do254} for aerospace, cover some aspects of the \ac{AIA} requirements for high-risk AI systems. However, these standards do not, in their current state, address \ac{AI} specific considerations for safety, transparency, and human oversight. Although efforts are being made to develop new safety standards for AI systems, their development and adherence are a time-intensive process. Quality models for \ac{AI} products, such as ISO/IEC 25059:2023 \cite{iso25059}, can help address the requirements set out in the \ac{AIA} early in the development cycle. Additionally, they provide the flexibility to include the attributes that may not be safety relevant, but ensure better quality.

In addition to the effect on the \ac{AI} regulatory landscape, the \ac{AIA} introduces additional challenges to compliance when several stakeholders are involved. Safety-critical \ac{AI} products are typically part of complex global supply chains, where many suppliers interact to produce the final product. In the automotive industry, for example, it is uncommon for a single entity to be responsible for the development and integration of all vehicle components. In such scenarios, demonstrating compliance to the \ac{AIA} becomes an increasingly challenging and intricate task. It is clear that organizations will need tools and methodologies to address the requirements laid out by the regulation. Specifically, a systematic methodology that aids organizations in verifying compliance is required. To facilitate this, our work leverages product quality models to break down the \ac{AIA} requirements into verifiable properties. In the first phase, an extended quality model for \ac{AI} systems is derived using attributes that are relevant to the \ac{AIA}. Next, using this quality model, an approach to map the articles of the \ac{AIA} to quality attributes for \ac{AI} Systems is presented. Finally, to address the complexities arising from supply chain relationships, a contract-based approach for the derivation of technical requirements from quality attributes is proposed. This methodology is, to the best of our knowledge, the first systematic approach for deriving technical requirements at the stakeholder level from high-level \ac{AIA} requirements.  

The contributions are as follows: 

\begin{itemize}
    \item An extended quality model for safety-critical \ac{AI} systems, which covers relevant attributes for the EU AI Act; 
    \item A systematic approach for mapping the \ac{AIA} requirements to relevant quality attributes in the extended quality model;
    \item A contract-based approach for deriving verifiable technical requirements for the quality attributes; and finally,
    \item An exemplary use case for an automotive supply chain is presented to demonstrate the applicability of the proposed methodology.
\end{itemize}

\section{Background}
There is currently little work surrounding how the requirements laid out for high-risk \ac{AI} systems should be addressed. Many organizations seek to understand whether compliance with current regulations can assist in addressing the EU AI Act. Existing  standards do not fully cover the stringent requirements laid out in the \ac{AIA}, such as transparency, lawfulness, and fairness. Product quality models may help fill this gap, and can be more easily adapted to include properties that may not be safety relevant but which do contribute to quality. Existing quality standards, such as ISO/IEC 25010 Product Quality Standard \cite{iso25010} and ISO/IEC FCD 25012 Data Quality Model \cite{iso25012} do not address \ac{AI} specific attributes such as transparency, controllability, and intervenability. The Quality Model for \ac{AI} Products/Systems in ISO 25059:2023 \cite{iso25059} introduces some \ac{AI} specific attributes like functional adaptability, and robustness, however, it is still lacking in its coverage of attributes like transparency, monitorability, and data quality, among others. ISO/IEC 24028 -  \textit{overview of trustworthiness in artificial intelligence} highlights the need for new standards which incorporate AI specific quality attributes \cite{iso24028}. Given this, recent contributions have addressed the need for extended quality models for \ac{AI} systems.

The authors of \cite{siebertGuidelinesAssessingQualities2020} define a systematic process for deriving a quality model for ML systems. They formalize the derivation of quality attributes using a quality meta-model, enabling the modelling of different hierarchies of quality. From this meta-model, relevant entities are defined and categorized into corresponding views of an ML product, namely the model, data, infrastructure, environment, and system views. Relevant properties are then described for a selected use case, and a list of corresponding metrics is proposed. This systematic approach ensures a comprehensive coverage of ML-related quality properties, however, it may not be well suited for addressing the \ac{AIA}. Given the high-level nature of the \ac{AIA}, it is beneficial to address high-level properties of \ac{AI} products, which may apply to several levels of abstraction and stakeholder perspectives. 
In addition, alignment with existing standards is relevant for organizations wishing to understand their current coverage in their development practices. As such, the extended quality model proposed in this paper is based on an alignment to high-level product quality standards, and other existing safety standards. 

Aside from quality models, recent literature has emerged proposing different approaches for addressing the \ac{AIA}. Novelli et al. \cite{novelli2023taking} highlight the importance of accurately assessing the risk of \ac{AI} systems in the context of the \ac{AIA}. The authors introduce a risk-assessment model to improve the accuracy of this risk estimation for ethical and safe \ac{AI} practices in accordance with the \ac{AIA}. While relevant to addressing the \ac{AIA}, this work focuses only on the risk classification and does not delve into the requirements for \ac{AI} systems that are deemed high risk. A different perspective is taken in \cite{sovrano2022metrics}, which provides an overview of explainability requirements in the \ac{AIA}, proposing metrics for assessing AI Act compliance. The authors highlight the need for metrics that are risk-focused, model-agnostic, goal-aware, intelligible, and accessible, and assess current metrics against these criteria. The paper provides a thorough coverage of explainability, but does not address the broader spectrum of requirements outlined in the EU AI Act. It also lacks a comprehensive methodology for addressing these requirements from the perspective of different stakeholders, leaving a gap in practical guidance for entities seeking compliance. A more pragmatic approach to compliance is suggested in \cite{walters2023complying}, where the authors propose a methodology for organizations to measure their compliance to the \ac{AIA} using a comprehensive questionnaire. However, the approach focuses on measuring compliance to the \ac{AIA}, and does not provide guidance to organizations who may seek further compliance.

\section{Proposed Methodology: Eliciting high-level requirements from the \ac{AIA}}
This section presents the systematic methodology for eliciting high-level requirements from the EU AI Act. First, an overview of the extended quality model for \ac{AI} Products is presented, followed by the approach for mapping \ac{AIA} requirements to quality attributes. Finally, a contract-based approach for deriving technical requirements for quality attributes is proposed. 

\subsection{Deriving an Extended Quality Model for safety-critical AI Systems}
\begin{figure*}[htbp!]
    \includegraphics[width = \textwidth]{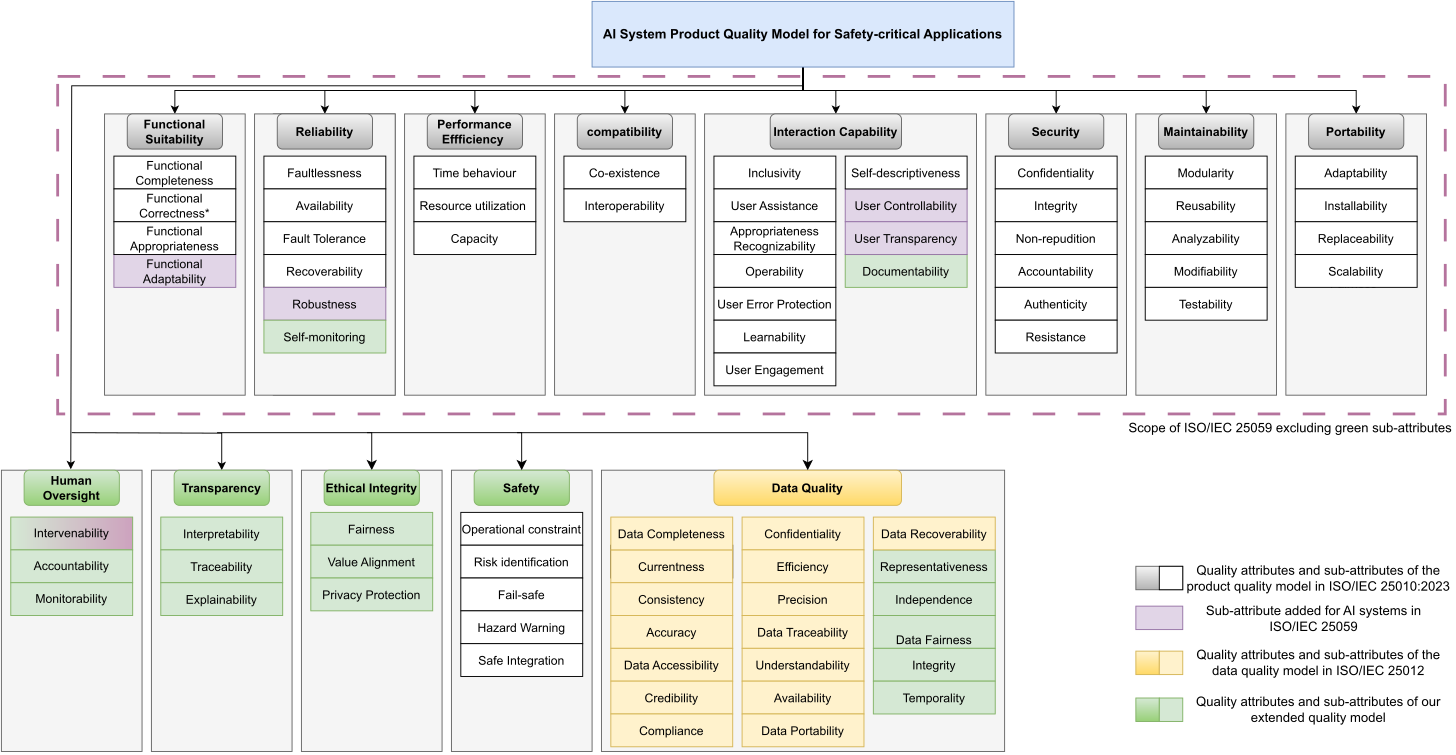}
    \caption{Extended Quality Model for AI products for safety-critical applications.}
    \label{fig:quality_model}
\end{figure*}

 To derive relevant quality attributes for safety-critical \ac{AI} systems, ISO/IEC 25059 \cite{iso25059} is used as a baseline. ISO/IEC 25059 provides the quality model serving as an extension to the ISO 25010:2011 series - Systems and Software Quality Requirements and Evaluation (SQuaRE)\cite{iso25010}. ISO/IEC 25059 defines quality attributes and sub-attributes that establish consistent terminology for specifying, measuring, and evaluating the quality of AI systems. It considers the quality model from two perspectives, product quality and quality
in use. In this report, we will only focus on the product quality model. The product quality model from ISO/IEC 25059 is highlighted in Figure~\ref{fig:quality_model}. 

We extend the product quality model presented in ISO/IEC 25059, with a specific focus on the following points:  
\begin{itemize}
    \item Covering relevant topics from the \ac{AIA} to increase trustworthiness. ISO/IEC 25059 has some gaps when it comes to the coverage of \ac{AIA} requirements, for example, there is a lack of consideration for human oversight, transparency for different stakeholders, and ethical integrity. We have added them as attributes in the extended quality model.   
    \item Integrating safety and data quality attributes in the ISO/IEC 25059 product quality model. The safety attribute, present in ISO/IEC 25010:2011 upon which ISO/IEC 25059 is based, is notably absent in ISO/IEC 25059. Similarly, the data quality model is extended from ISO/IEC 25012:2008 \cite{iso25012}. We have included it in our extended quality model due to the high dependence of the quality (including safety) of the AI systems on the quality of data.
    \item Incorporating AI-related safety properties and data quality from other sources, such as work from \cite{siebertConstructionQualityModel2022}, or the upcoming safety standard for AI systems in road vehicles, ISO PAS 8800 \cite{iso8800}. 
    \item Aligning ISO/IEC 25059:2023 with the updated version of ISO/IEC 25010:2023. It is currently based on ISO/IEC 25010:2011.
\end{itemize}
The extended model is depicted in Fig.~\ref{fig:quality_model}. Definitions for quality attributes and sub-attributes are given in Table~\ref{tab:qualityAttrDef}. This methodology can be adapted as new standards emerge regarding AI system product quality. For the safety characteristic, we recommend using domain-specific standards, where available, for a more systematic approach. For instance, combining ISO 26262 \cite{iso26262}, ISO 21448 \cite{iso21448}, and the upcoming ISO PAS 8800 \cite{iso8800} for road vehicles.

\begin{table}
\caption{New or Modified Definitions of Quality (Sub-)Attributes in the Extended Quality Model (see Figure~\ref{fig:quality_model}).}
\centering
\begin{tabular}{|p{1.8cm}|p{6.2cm}|} \hline
\textbf{Term} & \textbf{Definition} \\ \hline \hline
Ethical Integrity & The extent to which an entity's actions, beliefs, methods, measures, and principles all derive from a single core group of values. \\ \hline 
Human Oversight & The ability for humans to understand, supervise, and control the design and operation of AI-based systems. \cite{europeancommissionEthicsDesignEthics2021} \\ \hline 
Fairness & The extent to which a system prevents unjust predictions towards protected attributes (race, gender, income, etc). Ability of the model to output fair decisions. \cite{siebertGuidelinesAssessingQualities2020} \\ \hline 
Privacy \newline protection & The extent to which the product or system protects the privacy and handles sensitive information of the stakeholders involved (users, people in training examples). \\ \hline 
Value Alignment & The extent to which the AI system behaviour is aligned with human values. \cite{iso8800}  \\ \hline 
Self-Monitoring & The extent to which the system is aware of its state so it can respond appropriately to avoid going to a harmful state. \\ \hline 
Documentability & see ISO/IEC/IEEE 24765 \cite{IEEE24765} \\ \hline 
User Transparency & Degree to which the functionalities of the system are clear to the intended user.  \\ \hline
Interpretability & The extent to which the inner workings of the AI system can be analyzed in order to understand why it behaves the way it does. \\ \hline 
Traceability & The extent to which there exists data and processes that can record the system’s decisions and link artifacts at different stages. \cite{liQualityAttributesTrustworthy2023} \\ \hline 
Explainability & see ISO 22989 \cite{iso22989} \\ \hline 
Accountability & Capability of a product to enable actions of a human to be traced uniquely to the human. \\ \hline 
Monitorability & The extent to which relevant indicators of an AI system are effectively observed/monitored and integrated in the operation of the system . \\ \hline
Representative-\newline ness & The distribution of data (or probability of distribution) truly corresponds to the information in the environment or the phenomenon to be captured. \cite{iso8800}  \\ \hline 
Independence & The data at a specific level of architectural abstraction are not affected by changes to lower levels of abstraction. separate sets of data are used for specific purposes where required (e.g. AI training data, AI validation data).\cite{iso8800}  \\ \hline 
Data Fairness & Degree to which the data is free from bias against a given group. \cite{siebertGuidelinesAssessingQualities2020} \\ \hline 
Availability & The degree to which data has attributes that enable it to be retrieved by authorized users and/or applications in a specific context of use and within the time required. (see \cite{iso25012} and \cite{iso8800}) \\ \hline 
Integrity & The data are unaltered either by natural phenomenon (e.g. noise) or intentional action (e.g. poisoning). \cite{iso8800}  \\ \hline 
Temporality & A general property referring to temporal characteristics of data e.g. its timeliness, ageing or lifetime. \cite{iso8800}  \\
\hline
\end{tabular}\label{tab:qualityAttrDef}
\end{table}

\subsection{Mapping EU AI Act Articles to the Extended Quality Model}
The \ac{AIA} articles for high-risk \ac{AI} systems do not provide guidelines for achieving compliance. To enhance clarity, we propose to map these articles to our extended quality model. Such a mapping can be leveraged to assess the coverage of the \ac{AIA} based on measurable properties of \ac{AI} systems. We used our own experiences and research, coming from diverse research backgrounds, to consolidate a detailed mapping of high-level requirements to quality attributes. A high-level summary of the mapping is shown in Table \ref{table:mapping}. Using the mapping of the \ac{AIA} articles to quality attributes, relevant sub-attributes can be selected and verified using the contract-based approach proposed in the next section. 

\begin{table}[htbp!]
\caption{Mapping of EU AI Act Requirements to Quality Attributes for Safety-critical AI Systems.}
\centering
\begin{tabular}{|p{3cm} |p{5cm}|} \hline    
\textbf{Article} & \textbf{Sub-Attribute Mapping}  \\ \hline \hline
\textbf{9. Risk Management System}  & Risk identification, Testability, Value Alignment \\ \hline 
\textbf{10. Data and data governance}  & Independence, Data Completeness, Currentness, Independence, Data Fairness, Precision, Representativeness, Consistency, Accuracy, Credibility, Temporality, Confidentiality, Compliance, Data Traceability  \\ \hline 
\textbf{11. Technical Documentation}   &  Traceability \\ \hline 
\textbf{12. Record-keeping}   & Operability, Non-repudition, Traceability, Self-descriptiveness, Accountability, Self-Monitoring, User Engagement, Monitorability \\ \hline 
\textbf{13. Transparency and provision of information to users}  & User Engagement, Self-descriptiveness, User Transparency, Interpretability, Documentability, Appropiateness Recognizability \\ \hline 
\textbf{14. Human Oversight}   & Documentability, Learnability, Value Alignment, Accountability, Interpretability, Fairness, Explainability, Intervenability, Monitorability, User Error Protection. \\ \hline 
\textbf{15. Accuracy, robustness, and cybersecurity}   & Functional Correctness, Faultlessness, Robustness, Appropiateness Recognizability, Self-descriptiveness, Functional Adaptability, Fault Tolerance, Robustness, Integrity, Resistance \\ \hline
\end{tabular}\label{table:mapping}
\end{table}

\subsection{Contract-Based Validation Approach for Quality Attributes}
High-risk \ac{AI} applications are typically part of complex global supply chains, in which several stakeholders are involved. In this context, ensuring the fair, lawful, and ethical development of \ac{AI} applications is notably challenging. Parallels can be drawn with the recently enacted Supply Chain Act for companies headquartered in Germany \cite{SupplyChainAct}. This legislation extends the responsibility of organizations and mandates the safeguarding of human rights and environmental protection throughout the entire supply chain. While not specific to \ac{AI}, this legislation provides insights into how a company's responsibility for regulatory adherence is not simple, and may in some cases include indirect suppliers. A similar perspective can be applied to the \ac{AIA} where the responsibility is defined for some actors within the AI value chain, yet remains unspecified for others.

The \ac{AIA} defines a set of relevant \ac{AI} actors, and outlines responsibilities for compliance depending on these defined roles. Principal responsibility for compliance is assigned to the provider of a high-risk \ac{AI} system. However, in the case of safety-critical systems, any manufacturer in the supply chain can also be assigned responsibility. Importers and distributors are required to verify that a provider has met their obligations. End users, on the other hand, are mostly given rights in the framework of the Act, but proposals for amendments have been made to impose more requirements on them. Given the complexities arising from an ambiguous assignment of responsibilities, stakeholders will likely need to ensure not only their own compliance, but in certain cases the compliance of other involved actors. 

One of the few approaches to deriving a use-case agnostic, stakeholder-specific approach to compliance is provided in the EU Model contractual clauses for the procurement of \ac{AI} systems from external stakeholders. The clauses are generic and adaptable to specific use cases, and provide organizations wishing to procure \ac{AI} systems with a possible solution to ensuring compliance with the \ac{AIA}. The clauses are aligned with the \ac{AIA}, and support the ethical, transparent, and accountable development of \ac{AI} \cite{AIclauses}. The Commission highlights that these clauses may need to be adjusted depending on the contractual relationships. These clauses are thus limited in the sense that they do not distinguish between the obligations of the many actors discussed in the \ac{AIA}. Additionally, there is a need for a concrete methodology to derive technical requirements from such contractual clauses. We propose a contract-based approach for the systematic validation of the \ac{AIA} requirements across the value chain.

\begin{figure*}[htbp!]
\centering
    \includegraphics[width = 0.9\textwidth]{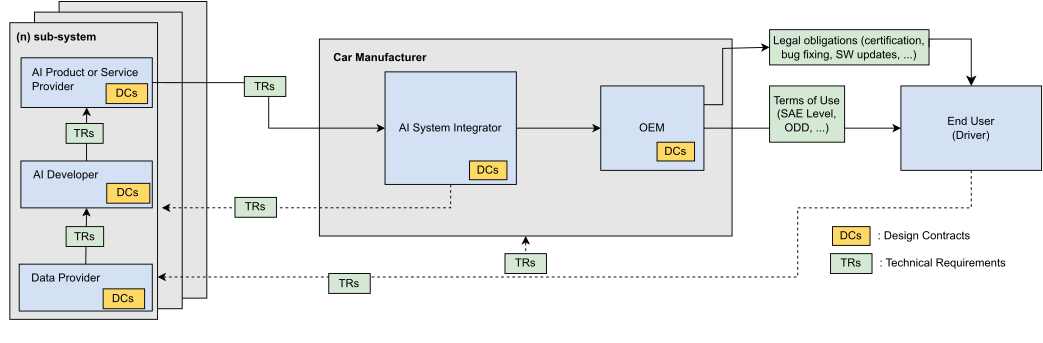}
    \caption{Automotive supply chain demonstrating stakeholder interactions and respective design contracts (DCs) and technical requirements (TRs).}
    \label{fig:supply chain}
\end{figure*}

Our approach is based on design contracts. Design contracts define \textit{guarantees} which are guaranteed to be fulfilled by the stakeholder. The fulfillment of said design contract is only guaranteed given that a set of \textit{assumptions} is fulfilled \cite{benvenisteContractsSystemDesign2018}. Verifying EU AI Act compliance thus boils down to the interface with the design contracts. Given that all \textit{assumptions} are fulfilled, \textit{guarantees} are assumed to be fulfilled. We demonstrate this approach using an exemplary automotive supply chain use case, shown in Fig. \ref{fig:supply chain}. For the sake of simplicity, we consider a car manufacturer which integrates (n) sub-systems. Each stakeholder in the supply chain may come from different entities. The design contracts (yellow boxes) are shown for each relevant stakeholder. The technical requirements (green boxes) are derived from the \textit{assumptions} and flow between stakeholders. Stakeholder definitions are taken from \cite{andradeArtificialIntelligenceAct2022} and \cite{iso22989}.  An example of this validation approach for a chosen quality attribute is presented in the following section.

\section{Use-Case Demonstration: Automotive Supply Chain}

\begin{figure}[htbp!]
\centering
    \includegraphics[width = 0.25\textwidth]{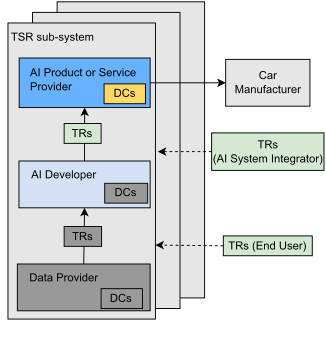}
\caption{Design contracts and technical requirements elicited by the AI Product or Service Provider.}
\label{fig:contracts}
\end{figure}

To demonstrate the applicability of our contract-based validation approach, we consider the typical automotive supply chain presented in Fig. \ref{fig:supply chain}. Suppose we have a \ac{TSR} component as a sub-system for a car manufacturer, as depicted in Fig. \ref{fig:contracts}. We would like to verify compliance for a given quality attribute in Table \ref{table:mapping}. Due to its applicability to Article 14: Transparency and Provision of Information to Users and Article 15: Human Oversight, we select Explainability (for definition see Table \ref{tab:qualityAttrDef}) as our quality sub-attribute. Starting with the \ac{AI} Product or Service Provider as our primary stakeholder, we would see stakeholder interactions as illustrated in Fig. \ref{fig:contracts}.

We first specify the design contract for the primary stakeholder. We define the \textit{assumptions} that are required so that the primary stakeholder can demonstrate compliance to the \ac{AIA}. In this case, we define \textit{assumptions} for explainability of the \ac{AI} component. These \textit{assumptions} are exemplary and would be refined depending on the relevant use case. From these \textit{assumptions}, the \ac{AI} product or service provider would define \textit{guarantees} that it can satisfy, given that the \textit{assumptions} are met. An example of a design contract for the AI Product or Service Provider is shown in Table \ref{tab: designcont}. 

\begin{table}[htbp!]
    \centering
    \caption{Design Contract for the AI Product or Service Provider.}
    \label{tab: designcont}
    \begin{tabular}{|c| p{0.8\linewidth}|}
        \hline
        \multicolumn{2}{|c|}{\textbf{\textit{Assumptions}}}  \\
        \hline \hline
        1 & The TSR component can be analyzed to understand its behavior. Documentation with global class-wise explanations is provided and representative. \\
        \hline
        2 & Appropriate documentation regarding the development of the TSR model is available. \\
         \hline
        3 & The TSR can express important factors influencing its predictions in a way that humans can understand. \\
         \hline
        4 & Documentation from the AI system integrator regarding how sub-systems interact in the overall car is available. \\
        \hline \hline
        \multicolumn{2}{|c|}{\textbf{\textit{Guarantees}}}  \\
         \hline \hline
         1 & Appropriate documentation regarding the design, development, licensing, and usage restrictions of the TSR is available. \\
        \hline
    \end{tabular}
\end{table}

The AI Provider's \textit{assumptions} would be detailed as technical requirements for the relevant stakeholders. In Table \ref{tab:techreq}, we provide examples of how these requirements might be formulated from the technical point of view in a legal contract. This approach provides a formal method to derive technical requirements for \ac{AIA} requirements using contract-based design. 

\begin{table}[htbp!]
\caption{Requirements given by the AI Product or Service Provider.}
\label{tab:techreq}
\begin{center}
\begin{tabular}{|p{1.5cm}|p{5cm}|p{1cm}|}
  \hline
 Technical Requirement & Description & Owner \\
  \hline \hline
    TR1 & The model architecture is well documented so that an expert user can understand the inner workings of the TSR component. & AI Developer \\ \hline 
    TR2 & An ex-ante explanation is available for the user of the AI system. For example, documentation containing global class-wise explanations is provided, using a state-of-the-art explainability method. & AI Developer \\ \hline 
    TR3 & Documentation containing train/test/validation data, pre- and post- processing operations, optimization method, loss function, and hyperparamaters used for training, is available. & AI Developer\\ \hline 
    TR4 & An ex-post explanation is available for the user of the AI system which satisfies the required level of explainability. For example, a local, post-modelling explainability method such as SHAP is implemented. & AI Developer \\ \hline 
    TR5 & The AI system integrator shall provide requirements for the TSR interface within the system.  & AI System Integrator \\ \hline
\end{tabular}
\end{center}
\end{table}

\subsection{Discussion}
This work describes a systematic methodology that can be used to assess the \ac{AIA} requirements from the perspective of different stakeholders. 
The proposed approach does not claim complete coverage of the \ac{AIA}. Instead, the extended quality model and the mapping should be subject to iterative refinement. This allows for continuous improvement as new insights emerge, regulatory frameworks evolve, or additional AI-specific attributes are identified or modified. 

The mapping does not provide a measure of the degree of coverage of each article. 
The goal of the mapping at this stage is to highlight the utility of quality models for addressing properties of AI models not addressed by current standards. Extensions to both our model and our methodology are expected in future work. 

The quantification of quality attributes remains a challenge. Current models lack precise metrics for evaluating critical aspects such as fairness, transparency, and adaptability in AI systems. This lack of metrics is particularly problematic in the context of contractual agreements, where clear and quantifiable measures are essential.

The practical implementation of certain quality attributes, such as human oversight, raises questions about the applicability of these requirements in real-world scenarios. In fully autonomous vehicles, the concept of oversight is unclear, necessitating a rethinking of how such systems are evaluated and regulated.

Additionally, while certain attributes were adequately defined for conventional software, their application to AI systems reveals new complexity. `Faultlessness' in AI, for instance, must consider the probabilistic nature of AI decisions, necessitating a redefinition that accounts for AI-specific error types and learning biases. This reassessment is crucial for ensuring that the extended model not only introduces new attributes for AI but also appropriately reinterprets existing ones to align with the unique characteristics and demands of AI technologies.

\section{Conclusion}
The EU AI Act is a transformative legislation which reshapes the global landscape of fair and ethical \ac{AI} development. In this paper, we present a systematic methodology for addressing the requirements for high-risk AI products introduced in the \ac{AIA}. We develop an extended quality model for \ac{AI} systems, and propose to map these quality attributes to the \ac{AIA} requirements. To address compliance, a contract-based approach for defining technical requirements is presented, ensuring that stakeholders across complex supply chains adhere to the EU AI \ac{AIA} regulations. Our design contracts foster a flexible and structured approach to compliance. This methodology allows researchers and practitioners to bridge the gap between existing quality models and the regulatory demands of the \ac{AIA}. This facilitates the development and assessment of \ac{AI} systems that not only adhere to established quality standards but also comply with the regulatory requirements outlined in the \ac{AIA}.

\bibliographystyle{IEEEtran}

\end{document}